\documentclass[lettersize,journal]{IEEEtran}
\usepackage{amsmath,amsfonts}
\usepackage{algorithmic}
\usepackage{algorithm}
\usepackage{array}
\usepackage[caption=false,font=normalsize,labelfont=sf,textfont=sf]{subfig}
\usepackage{textcomp}
\usepackage{stfloats}
\usepackage{url}
\usepackage{verbatim}
\usepackage{graphicx}
\usepackage{cite}
\usepackage{booktabs}
\usepackage{multirow}
\usepackage{makecell}
\usepackage{pifont}

\usepackage[colorlinks=true,citecolor=green,linkcolor=red,urlcolor=blue]{hyperref}

\usepackage{makecell}

\begin{document}

\title{CLUIE: Clustering-Aware Recurrent Propagation with Local Structural Compensation for Underwater Image Enhancement}

 \author{Kui~Jiang,~\IEEEmembership{Member,~IEEE,}
         Zefan~Feng,
         Laibin~Chang,
         Yan~Luo,\\
         Junjun~Jiang,~\IEEEmembership{Senior Member,~IEEE,}
         Xiaopeng~Fan,~\IEEEmembership{Senior Member,~IEEE}
        
\thanks{This research was financially supported by the National Natural Science Foundation of China (62501189), the Natural Science Foundation of Heilongjiang Province of China for Excellent Youth Project (YQ2024F006).} 
\thanks{Kui~Jiang, Zefan~Feng, Yan~Luo, Junjun~Jiang and Xiaopeng~Fan are with the School of Computer Science and Technology, Harbin Institute of Technology, Harbin, China (e-mail: jiangkui@hit.edu.cn).}
\thanks{Laibin~Chang is with the School of Computer Science, Wuhan University, Wuhan, China (changlb666@whu.edu.cn).}
}

\markboth{IEEE Transactions on Image Processing}%
{Kui \MakeLowercase{\textit{et al.}}: Clustering-Aware Recurrent Propagation with Local Structural Compensation for Underwater Image Enhancement}



\maketitle

\begin{abstract}

Underwater image enhancement is challenging due to wavelength-dependent absorption, scattering, and backscattering, which jointly cause color distortion, contrast degradation, and detail loss. Moreover, underwater degradation is often spatially heterogeneous: different regions within an image may suffer from different degradation patterns, whereas spatially distant regions may share similar restoration requirements. This calls for enhancement models that can preserve local structures while adaptively modeling long-range dependencies across content-related regions. 
Although visual Receptance Weighted Key Value models provide an efficient linear-complexity solution for long-range dependency modeling, their recurrent state propagation typically follows predefined scanning orders. Such content-agnostic trajectories are not well aligned with the spatially non-uniform degradation characteristics of underwater images. To overcome this limitation, we propose CLUIE, a clustering-aware recurrent trajectory modeling framework for underwater image enhancement. The core idea is to replace fixed recurrent propagation paths with content-adaptive token trajectories. To this end, we design Clustering-aware Semantic Dynamic Reordering (CSDR), which clusters tokens in the learned feature space and derives recurrent traversal orders according to inter-cluster contextual relationships. As a result, WKV states are encouraged to propagate among feature-related regions rather than along rigid spatial or spectral scans. 
To compensate for the potential loss of local spatial continuity introduced by dynamic reordering, we further propose Dark-response Modulated Local Propagation (DMLP). DMLP extracts local structural responses using depth-wise convolution and adaptively modulates their injection through a feature-space pseudo-dark response statistic. By coupling content-adaptive recurrent aggregation with locally modulated structural compensation, CLUIE effectively addresses spatially heterogeneous underwater degradation. Extensive experiments on multiple underwater image enhancement benchmarks show that CLUIE achieves competitive quantitative performance, favorable computational efficiency, and visually balanced restoration quality.Code is available at:https://github.com/geekpool/CLUIE

\end{abstract}


\section{Introduction}
\IEEEPARstart{U}{nderwater} Image Enhancement (UIE) aims to recover visually clear, color-consistent, and structurally faithful images from degraded underwater observations~\cite{9788535,li2021underwater,7574330}. It is an important preprocessing technique for underwater robotics, marine ecological monitoring, visual inspection, and underwater object perception. However, underwater imaging is affected by wavelength-dependent light absorption, scattering, backscattering, and non-uniform illumination. These degradation factors usually lead to severe color casts, low contrast, haze-like veiling effects, and loss of fine details~\cite{8954437}.

\begin{figure}[t]
\centering
\includegraphics[width=\columnwidth]{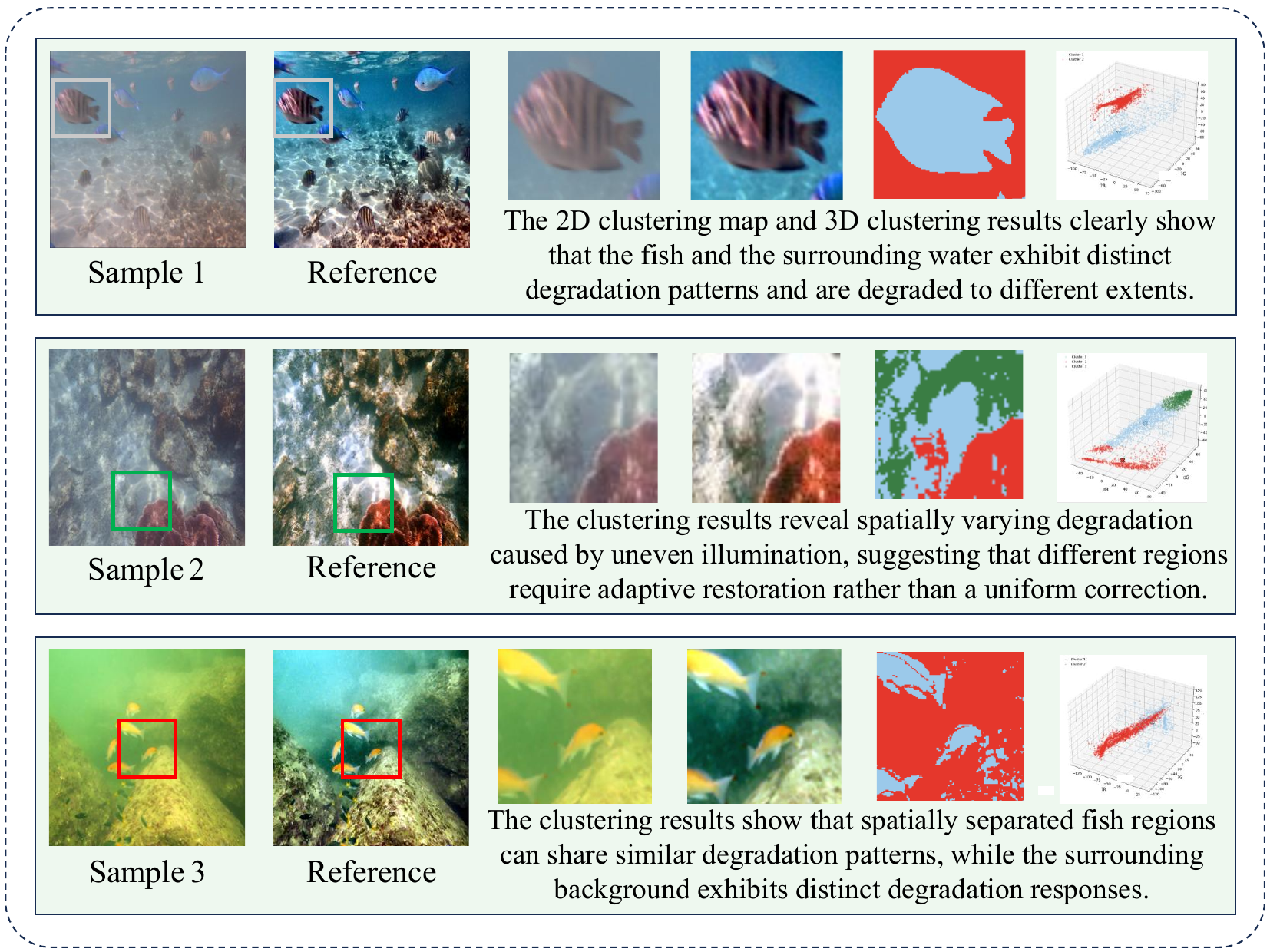}\vspace{-2mm}
\caption{Illustration of spatially non-uniform underwater degradation based on signed RGB-change clustering. Pixel-wise $\Delta \mathbf{RGB} = (\Delta R, \Delta G, \Delta B)$ between degraded images and references is clustered to obtain 2D degradation maps and 3D RGB-change distributions.}\vspace{-2mm}
\label{fig:intro_nonuniform}
\end{figure}

Different from many general image restoration tasks, underwater degradation is often spatially heterogeneous. The degradation strength and color shift vary with scene depth, water medium, illumination condition, and object reflectance. As a result, different regions in the same image may require different restoration operations. As shown in Fig.~\ref{fig:intro_nonuniform}, foreground objects may need color correction and structure preservation, while background water regions may require haze suppression and contrast enhancement. Moreover, spatially distant regions may share similar degradation responses, whereas adjacent regions may exhibit different degradation characteristics. This observation suggests that UIE should not be treated as a purely global color correction problem or a purely local detail enhancement problem. Instead, it requires restoration models that can adaptively organize dependencies according to image content and degradation-related feature correlations.

Deep learning has significantly advanced UIE through convolutional and attention-based architectures. CNN-based methods exploit local receptive fields and hierarchical representations to improve color correction, contrast enhancement, and detail reconstruction~\cite{li2020underwater,10048777}. However, their dependency modeling ability is mainly constrained by local receptive fields, making it difficult to associate spatially distant regions with similar degradation patterns. Transformer-based restoration models introduce self-attention to model long-range dependencies \cite{NIPS2017_3f5ee243,liang2021swinir,wang2022uformer,Zamir2022Restormer}, but their quadratic complexity with respect to the number of tokens limits efficiency for high-resolution images. In addition, many Transformer variants organize dependencies through fixed windows, hierarchical partitions, or predefined attention layouts, which may not explicitly match the region-dependent degradation characteristics of underwater scenes.


\begin{figure*}[t]
\centering
\includegraphics[width=\textwidth]{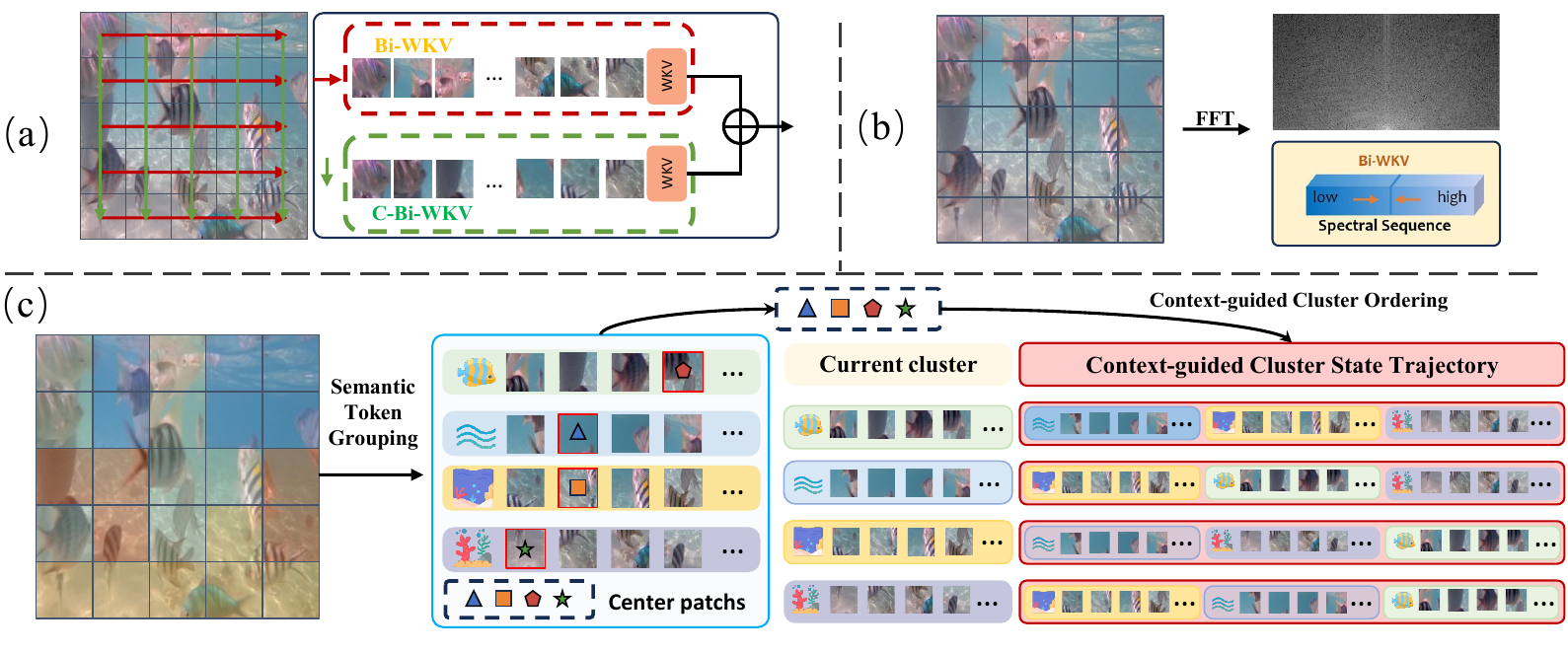}\vspace{-2mm}
\caption{Comparison of sequence organization strategies in visual RWKV. 
(a) RWKV-IR builds dependencies through horizontal and vertical spatial scans. 
(b) Fourier-RWKV performs recurrent aggregation in a fixed spectral order.
(c) Our method constructs a content-adaptive recurrent trajectory by semantic clustering and intra-cluster spatial reordering, allowing token propagation to follow semantic and degradation-related correlations.}\vspace{-4mm}
\label{fig:scan_compare}
\end{figure*}

Recently, state-space~\cite{gu2023mamba,liu2024vmamba,guo2024mambair} and recurrent models have provided promising alternatives for efficient long-range modeling. In particular, Receptance Weighted Key Value (RWKV)~\cite{Peng2023RWKV} combines recurrent state propagation with Transformer-like parallel training and linear-time inference, making it attractive for visual restoration tasks. When RWKV is applied to images, two-dimensional feature maps need to be serialized into one-dimensional token sequences before WKV aggregation. Therefore, the token order directly determines the recurrent state propagation path. Existing visual RWKV methods usually adopt predefined traversal strategies. For example, spatial RWKV variants employ horizontal, vertical, or multi-directional scans to preserve two-dimensional spatial structures~\cite{du2024exploring}, while frequency-domain RWKV methods perform recurrent aggregation according to fixed spectral orders~\cite{zheng2025fourier}. These designs improve visual sequence modeling from spatial or spectral perspectives, but their traversal paths are still independent of the input image content.

Such content-agnostic scanning strategies are not always suitable for underwater image enhancement. In underwater scenes, restoration-related dependencies may exist between non-adjacent regions with similar semantic or degradation-related feature responses. Conversely, neighboring regions may suffer from different degradation effects and require different enhancement operations. As illustrated in Fig.~\ref{fig:scan_compare}, a fixed raster, directional, or spectral traversal may therefore mix restoration-irrelevant tokens and delay interactions between regions that should be jointly modeled. This motivates us to revisit a fundamental yet under-explored problem in visual RWKV-based restoration: how should two-dimensional visual tokens be organized into a one-dimensional recurrent trajectory for spatially heterogeneous degradation?


To answer this question, we propose CLUIE, a content-adaptive recurrent trajectory modeling framework for underwater image enhancement. The key idea is to reformulate fixed RWKV token serialization as input-dependent recurrent trajectory construction. Instead of forcing WKV states to propagate along a predefined spatial or spectral path, CLUIE dynamically reorders tokens according to feature-level contextual relations. In this way, recurrent aggregation can be better aligned with content-related and degradation-related dependencies in underwater images while preserving the linear modeling advantage of RWKV.

The core component of CLUIE is a Clustering-aware Recurrent RWKV block, termed CRWKV. Each CRWKV block consists of two complementary modules: Clustering-aware Semantic Dynamic Reordering (CSDR) and Dark-response Modulated Local Propagation (DMLP). CSDR first groups tokens in the learned feature space and then constructs a cluster-level traversal path according to inter-cluster contextual relations. To avoid excessive spatial discontinuity, tokens within each cluster are further traversed in a spatially coherent manner. The resulting trajectory is used to reorder tokens before WKV aggregation, while the output is finally mapped back to the original spatial layout. Therefore, CSDR changes the recurrent state propagation order without altering the spatial correspondence of the restored feature map.

Although dynamic token reordering improves content-adaptive long-range aggregation, it may weaken the local continuity of original image neighborhoods. Local continuity is essential for edge preservation, texture recovery, and fine detail reconstruction in UIE. To compensate for this effect, we introduce DMLP. DMLP extracts local structural responses using depth-wise convolution with layer-dependent receptive fields. It further computes a feature-space pseudo-dark response statistic to modulate the injection strength of local responses. This pseudo-dark response is not used as a physical transmission estimate; instead, it serves as a spatial modulation cue inspired by low-response feature statistics. By integrating DMLP before recurrent aggregation, CLUIE preserves local structural cues while benefiting from content-adaptive long-range propagation.

The main contributions are summarized as follows:
\begin{itemize}
    \item We propose CLUIE, a content-adaptive recurrent trajectory modeling framework for underwater image enhancement. It revisits the sequence organization problem of visual RWKV and adapts recurrent state propagation to spatially heterogeneous underwater degradation.
    
    \item We develop Clustering-aware Semantic Dynamic Reordering (CSDR), which constructs input-dependent RWKV propagation paths by feature-space token grouping, inter-cluster contextual ordering, and spatially coherent intra-cluster traversal. This enables recurrent states to propagate along content-related regions rather than fixed spatial or spectral scans.
    
    \item We introduce Dark-response Modulated Local Propagation (DMLP) to compensate for the local continuity loss introduced by dynamic token reordering. DMLP combines depth-wise local structural response extraction with feature-space pseudo-dark-response modulation, improving edge preservation and local detail recovery.
\end{itemize}
    

    

\section{Related Work}
\subsection{Underwater Image Enhancement}
Underwater image enhancement (UIE) has been studied through both model-driven and data-driven approaches. Physical and prior-based methods estimate transmission, background light, attenuation, or backscattering according to underwater imaging assumptions \cite{8578801}. Representative works include Sea-Thru, which models backscatter and range-dependent attenuation \cite{8954437}, dark-channel-prior variants for underwater scattering removal \cite{he2010single}, and depth-, illumination-, or fusion-based enhancement methods \cite{6104148,zhou2021underwater,ancuti2012enhancing}. These methods are interpretable and require limited training data, but their performance depends on the validity of simplified imaging assumptions and handcrafted priors. In real underwater scenes with non-uniform illumination and coupled degradations, inaccurate parameter estimation may lead to unstable restoration.


Deep learning methods have become dominant in UIE because of their stronger representation learning capacity. CNN-based methods learn direct or residual mappings from degraded images to enhanced images and have been widely used for color correction, contrast enhancement, and detail reconstruction. General restoration architectures such as U-Net, ResNet, and multi-stage restoration networks provide effective local feature extraction and hierarchical reconstruction~\cite{he2016deep,zamir2021multi,chen2022simple}. In UIE, Water-Net introduces a benchmark dataset and learns confidence maps to fuse multiple enhanced inputs~\cite{li2019underwater}; UWCNN designs water-type-specific branches for underwater restoration~\cite{li2020underwater}; and subsequent methods further exploit transmission guidance, multi-color-space embedding, guided information flow, artificial-light modeling, correlation feature matching, and multi-branch degradation aggregation~\cite{zhou2023ugif,li2023uialn,qi2021underwater}. Although CNN-based methods are effective for local structure recovery, their dependency modeling is mainly constrained by local receptive fields, which limits their ability to associate spatially distant regions that may share similar degradation responses.


Transformer-based methods improve long-range dependency modeling through self-attention \cite{NIPS2017_3f5ee243,dosovitskiy2020image,Liu2021Swin}. Generic restoration models such as IPT, SwinIR, Uformer, and Restormer demonstrate the benefit of global-local interaction in image restoration \cite{chen2021pre,liang2021swinir,wang2022uformer,Zamir2022Restormer}. These methods are better suited for global context modeling than CNNs, but self-attention incurs considerable computational cost for high-resolution images. Moreover, their dependency organization is usually determined by fixed windows, hierarchical partitions, or predefined attention structures, rather than being explicitly adapted to region-dependent degradation relations.

Recently, state-space and recurrent models provide a promising alternative for efficient long-range modeling with linear complexity~\cite{gu2023mamba,liu2024vmamba}. MambaIR introduces state-space modeling into generic image restoration~\cite{guo2024mambair}, while WaterMamba, Mamba-UIE, and O-Mamba apply Mamba-like architectures to underwater image enhancement for efficient global context aggregation~\cite{guan2024watermamba,zhang2024mamba,dong2025mamba}. MaIR and EVSSM further improve locality preservation and efficient visual state-space modeling for image restoration~\cite{li2025mair, kong2025efficient}. These methods reduce the computational burden of long-range dependency modeling compared with attention-based models. However, they mainly focus on efficient information propagation, while the organization of restoration-related dependencies under spatially heterogeneous underwater degradation remains less explored. In contrast, our work focuses on constructing content-adaptive recurrent trajectories for UIE.


\subsection{RWKV for Visual Restoration}

Receptance-Weighted Key-Value (RWKV) is a recurrent sequence modeling architecture that combines recurrent computation with Transformer-like parallel training and linear-time inference~\cite{Peng2023RWKV}. Unlike self-attention, RWKV aggregates contextual states along a one-dimensional token sequence. Therefore, when RWKV is applied to image restoration, the serialization order of two-dimensional visual features directly affects the path of recurrent state propagation.

Existing visual RWKV methods mainly improve sequence organization through predefined spatial or spectral traversal. Vision-RWKV extends RWKV to visual representation learning with bidirectional WKV and multi-directional token shift~\cite{duan2025vision}. RWKV-IR introduces horizontal and vertical scans for image restoration, which better preserve two-dimensional spatial structure than naive raster ordering~\cite{du2024exploring}. Fourier-RWKV further performs recurrent aggregation in the frequency domain using a fixed low-to-high frequency order~\cite{zheng2025fourier}. Recent works such as Restore-RWKV and URWKV enhance local interaction or multi-state modeling to better support low-level vision tasks~\cite{yang2025restore,xu2025urwkv}.

Although these methods improve RWKV for visual restoration, their token orders remain predefined and content-agnostic. For underwater images, restoration-related dependencies are often not aligned with fixed spatial or frequency traversal: distant regions may share similar degradation patterns, whereas adjacent regions may require different restoration operations. Therefore, fixed scanning orders may not provide an appropriate propagation path for region-dependent UIE.

Unlike Mamba-based UIE methods, which mainly improve efficient sequence modeling for global context aggregation, our work focuses on how the recurrent path itself should be organized for spatially heterogeneous underwater degradation. Unlike RWKV-IR and Fourier-RWKV, which rely on fixed spatial or spectral orders, CLUIE constructs an input-dependent token trajectory through clustering-aware feature reordering. In addition, we introduce dark-response modulated local propagation to compensate for the local continuity loss caused by dynamic reordering. This design enables CLUIE to couple content-adaptive long-range recurrent aggregation with local structural compensation for underwater image enhancement.

\begin{figure*}[!tp]
    \centering
    \includegraphics[width=\textwidth]{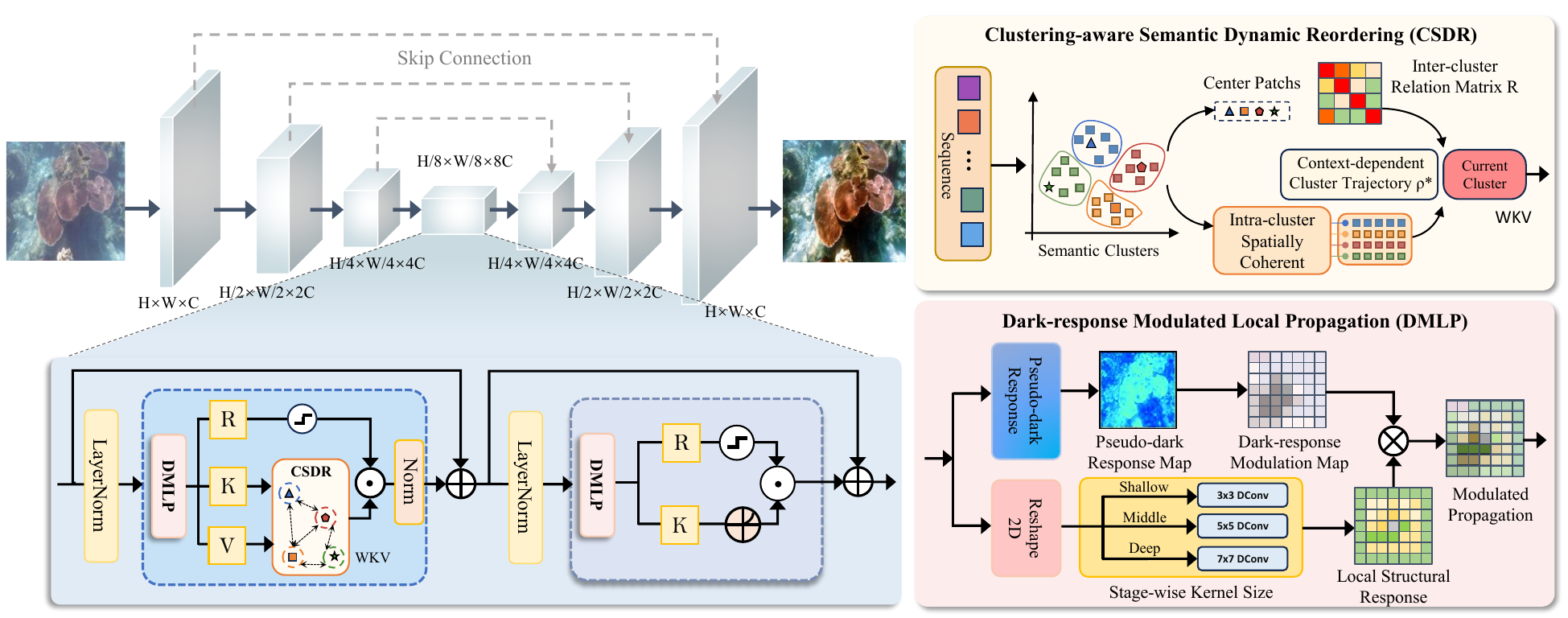}\vspace{-2mm}
    \caption{Overall architecture of the proposed CLUIE framework for underwater image enhancement. The model adopts an encoder-decoder structure built upon CRWKV blocks.Within each block, CSDR reconstructs the RWKV recurrent path into a content-adaptive token trajectory through semantic clustering and context-dependent reordering, while DMLP injects dark-response modulated local structural propagation to compensate for neighborhood continuity.}\vspace{-2mm}
    \label{fig:pipeline}
\end{figure*}

\section{Proposed Method}
\subsection{Overall Framework}
\label{subsec:overall}
CLUIE is designed to adapt RWKV state propagation to the spatially heterogeneous degradations in underwater images. In visual RWKV, two-dimensional features are serialized into a one-dimensional token sequence before WKV aggregation. Therefore, the token order determines the recurrent propagation path. Fixed raster, directional, or spectral scans provide efficient implementations, but their traversal rules are independent of image content and may not match region-dependent restoration relations.

As shown in Fig.~\ref{fig:pipeline}, CLUIE adopts an encoder-decoder architecture and uses the proposed CRWKV block as the basic modeling unit. Each CRWKV block contains two key components: Clustering-aware Semantic Dynamic Reordering (CSDR) and Dark-response Modulated Local Propagation (DMLP). CSDR constructs a content-adaptive token trajectory for WKV aggregation by grouping tokens in the feature space and reordering them according to inter-cluster contextual relations. DMLP injects modulated local structural cues before dynamic reordering to compensate for the potential loss of neighborhood continuity.

Given an input $\mathbf{X}$, one CRWKV block is formulated as 
\begin{equation}
\begin{aligned}
    \hat{\mathbf{X}} &= \mathrm{LN}(\mathbf{X}), \\
    \mathbf{X}_{loc} &= \mathrm{DMLP}(\hat{\mathbf{X}}), \\
    \mathbf{X}_{re} &= \mathrm{CSDR}(\mathbf{X}_{loc}), \\
    \mathbf{Y} &= \mathbf{X} + \mathrm{WKV}(\mathbf{X}_{re}),
\end{aligned}
\label{eq:overall_block}
\end{equation}
where $\mathrm{LN}(\cdot)$ denotes layer normalization and $\mathrm{WKV}(\cdot)$ denotes Receptance-Weighted Key-Value aggregation. The residual connection preserves the original representation, while CSDR only changes the intermediate token order used for recurrent aggregation. After WKV aggregation, the output is mapped back to the original spatial layout, ensuring that the input-output spatial correspondence is unchanged.

\subsection{Clustering-aware Semantic Dynamic Reordering}
\label{subsec:csdr}
In RWKV-based visual restoration, recurrent states are accumulated sequentially. Thus, the serialization order determines which tokens interact earlier and more directly. Existing visual RWKV methods usually adopt fixed spatial or spectral scans. Such orders are simple and efficient, but they assume a shared traversal rule for all images. This assumption is not always suitable for UIE, where distant regions may share similar degradation characteristics and adjacent regions may require different restoration operations. CSDR addresses this issue by constructing an input-dependent recurrent trajectory from the current feature map.


\subsubsection{Semantic Token Grouping}
Given a token feature $\mathbf{x}_t\in\mathbb{R}^{C}$ at position $t$, where $t\in\{1,\ldots,N\}$ and $N=H\times W$, we first normalize it as 
\begin{equation}
\mathbf{z}_t =
\frac{\mathbf{x}_t}{\|\mathbf{x}_t\|_2+\epsilon},
\label{eq:semantic_embedding}
\end{equation}
where $\epsilon$ is a small constant for numerical stability. The normalization reduces the influence of feature magnitude, which may vary with local illumination, contrast, and color degradation. The normalized tokens are then grouped by $K$-means clustering:

\begin{equation}
\{\mathcal{C}_k\}_{k=1}^{K}
=
\mathrm{KMeans}(\{\mathbf{z}_t\}_{t=1}^{N}),
\label{eq:kmeans_grouping}
\end{equation}
where $\mathcal{C}_k$ denotes the $k$-th token group. The grouping is performed in feature space and is not intended to produce explicit physical degradation regions. It provides a content-dependent partition for subsequent trajectory construction.

\subsubsection{Cluster-level Trajectory Construction}
For each cluster $\mathcal{C}_k$, we compute its semantic prototype and spatial centroid:
\begin{equation}
\mathbf{m}_k^{\mathrm{sem}}
=
\frac{1}{|\mathcal{C}_k|}
\sum_{t\in\mathcal{C}_k}\mathbf{x}_t,
\qquad
\mathbf{m}_k^{\mathrm{spa}}
=
\frac{1}{|\mathcal{C}_k|}
\sum_{t\in\mathcal{C}_k}\mathbf{p}_t,
\label{eq:cluster_prototype}
\end{equation}
where $\mathbf{p}_t\in[0,1]^2$ is the normalized coordinate of token $t$. The relation between clusters $\mathcal{C}_p$ and $\mathcal{C}_q$ is defined as

\begin{equation}
\begin{aligned}
\mathcal{R}_{p,q}
&=
\alpha\,\mathrm{CosSim}
\bigl(
\mathbf{m}_p^{\mathrm{sem}},
\mathbf{m}_q^{\mathrm{sem}}
\bigr)  \\
&\quad+
(1-\alpha)
\exp
\bigl(
-\|
\mathbf{m}_p^{\mathrm{spa}}
-
\mathbf{m}_q^{\mathrm{spa}}
\|_2
\bigr),
\end{aligned}
\label{eq:cluster_relation}
\end{equation}
where $\alpha$ balances feature affinity and spatial proximity. The feature-affinity term encourages related token groups to be visited close to each other, while the spatial-proximity term discourages excessively discontinuous transitions.

The cluster-level trajectory is denoted as 
\begin{equation}
\rho=[\rho_1,\rho_2,\ldots,\rho_K],
\label{eq:cluster_trajectory}
\end{equation}
where $\rho_i$ is the index of the $i$-th visited cluster. In implementation, $\rho$ is generated from the relation matrix $\mathbf{R}\in\mathbb{R}^{K\times K}$ using a greedy relation-guided traversal. Starting from the cluster with the largest average relation score, the next cluster is selected as the unvisited cluster with the highest relation to the current one. This yields an input-dependent cluster order without introducing additional learnable parameters.

\subsubsection{Spatially Coherent Intra-cluster Traversal}
Tokens within the same semantic cluster may be spatially disconnected. Directly traversing them as a single unordered set can introduce abrupt jumps and weaken local continuity. To reduce this effect, each cluster mask is projected back to the two-dimensional grid and split into connected components. Tokens inside each component are visited in raster order, and components within the same cluster are ordered by the raster indices of their spatial centroids.


Let $\Pi_k$ denote the intra-cluster sequence of cluster $\mathcal{C}_k$. Given the cluster-level order $\rho$, the final token trajectory is constructed as
\begin{equation}
\pi
=
\Pi_{\rho_1}
\oplus
\Pi_{\rho_2}
\oplus
\cdots
\oplus
\Pi_{\rho_K},
\label{eq:dynamic_token_trajectory}
\end{equation}
where $\oplus$ denotes sequence concatenation. This trajectory preserves local ordering within connected regions while allowing the global recurrent path to follow content-dependent cluster relations.

\subsubsection{Reordered WKV Aggregation}

Given the trajectory $\pi$, WKV aggregation is applied to the reordered sequence:
\begin{equation}
\mathbf{y}_{\pi,i}
=
\frac{
\sum_{j=1}^{i}
e^{-(i-j)\mathbf{w}+\mathbf{k}_{\pi,j}}
\odot
\mathbf{v}_{\pi,j}
}{
\sum_{j=1}^{i}
e^{-(i-j)\mathbf{w}+\mathbf{k}_{\pi,j}}
},
\label{eq:reordered_wkv}
\end{equation}
where $\mathbf{w}$ is the learnable decay vector, and $\mathbf{k}_{\pi,j}$ and $\mathbf{v}_{\pi,j}$ are the Key and Value features in the reordered sequence. All operations are channel-wise. The output is then mapped back to the original spatial layout:
\begin{equation}
\mathbf{y}=\pi^{-1}(\mathbf{y}_{\pi}).
\label{eq:inverse_permutation}
\end{equation}

Thus, CSDR modifies the recurrent state propagation order rather than the WKV operator itself. The spatial arrangement of the output remains consistent with the input feature map.

\subsection{Dark-response Modulated Local Propagation}
\label{subsec:dmlp}
CSDR improves long-range recurrent aggregation by dynamic reordering, but this operation may reduce the continuity of original spatial neighborhoods. Local continuity is important for edge preservation, texture recovery, and detail reconstruction in UIE. Existing visual RWKV methods often use Q-shift to introduce local information \cite{duan2025vision,du2024exploring}. However, Q-shift applies fixed channel-group shifts to all positions and does not adapt to local feature responses.

DMLP replaces fixed shifting with learnable and spatially modulated local propagation. It first extracts local structural responses using depth-wise convolution and then controls their injection strength with a feature-space pseudo-dark response map. The module is applied before the generation of Key, Value, and Receptance features.


\subsubsection{layer-based Local Response}
Given a feature map $\mathbf{F}\in\mathbb{R}^{B\times C\times H\times W}$, DMLP computes local responses by
\begin{equation}
\mathbf{L}
=
\mathrm{DConv}_{s_l}
\left(
\mathbf{F}
\right),
\label{eq:dmlp_local_response}
\end{equation}
where $\mathrm{DConv}_{s_l}(\cdot)$ is depth-wise convolution with window size $s_l$ at layer $l$. The window size is set as

\begin{equation}
s_l
=
\min(3+2l, 7).
\label{eq:dmlp_layer_window}
\end{equation}
Thus, shallow layers use smaller neighborhoods, while deeper layers aggregate broader local context. Since the convolution is depth-wise, it extracts channel-wise local structures without directly mixing channel semantics.


\subsubsection{Pseudo-dark Response Modulation}


To modulate local response injection, we compute a pseudo-dark response from intermediate features. For each spatial position $(u,v)$, the channel-minimum response is
\begin{equation}
\mathbf{d}_{F}(u,v)
=
\min_{c\in\{1,\ldots,C\}}
\mathbf{F}_{c}(u,v).
\label{eq:dmlp_channel_min}
\end{equation}

A local averaged statistic is then obtained as
\begin{equation}
\mathbf{D}_{F}(u,v)
=
\frac{1}{|\Omega_{s_d}(u,v)|}
\sum_{(m,n)\in\Omega_{s_d}(u,v)}
\mathbf{d}_{F}(m,n),
\label{eq:dmlp_pseudo_dark_response}
\end{equation}
where $\Omega_{s_d}(u,v)$ denotes an $s_d\times s_d$ neighborhood. We use average pooling instead of a second minimum operation to reduce sensitivity to isolated extreme responses.


The pseudo-dark response is computed from feature maps, not from RGB intensities. It is therefore used as a feature-space response statistic rather than a physical transmission estimate. After per-sample min-max normalization, the modulation map is defined as
\begin{equation}
\mathbf{A}
=
1
-
\mathrm{Norm}
\left(
\mathbf{D}_{F}
\right),
\label{eq:dmlp_pseudo_dark_modulation}
\end{equation}
where $\mathbf{A}\in\mathbb{R}^{B\times 1\times H\times W}$. The modulated local feature is obtained by residual injection:
\begin{equation}
\mathbf{F}^{loc}
=
\mathbf{F}
+
\lambda_l
\left(
\mathbf{A}
\odot
\mathbf{L}
\right),
\label{eq:dmlp_local_residual}
\end{equation}
where $\lambda_l$ is a learnable scaling factor. In our implementation, $s_d$ is fixed to $3$, while $s_l$ varies with layer depth.


\subsubsection{Branch-wise RWKV Integration}

The local feature $\mathbf{F}^{loc}$ is flattened back to tokens:
\begin{equation}
\mathbf{X}^{loc}
=\mathrm{Flatten}
\left(\mathbf{F}^{loc}\right)
\in\mathbb{R}^{B\times N\times C}.
\label{eq:dmlp_token_feature}
\end{equation}

\begin{table*}[!htb]
\caption{Full-reference quantitative comparison on UIEB, LSUI, and EUVP. The best, second-best, and third-best results are highlighted in \textcolor{red}{red}, \textcolor{blue}{blue}, and \textcolor{green}{green}, respectively. Avg. PSNR denotes the average PSNR over the three datasets. Model efficiency is evaluated by parameters, FLOPs, and inference time.}
\label{tab:full_ref}
\tabcolsep 0.050 in
\centering
\begin{tabular}{lc ccc ccc ccc c ccc}
\toprule
\multirow{2}{*}{\raisebox{-0.8ex}{Method}} & \multirow{2}{*}{\raisebox{-0.8ex}{Source}}
& \multicolumn{3}{c}{UIEB} & \multicolumn{3}{c}{LSUI} & \multicolumn{3}{c}{EUVP} & Avg. & Params & FLOPs & Time \\
\cmidrule(lr){3-5} \cmidrule(lr){6-8} \cmidrule(lr){9-11}
& & PSNR $\uparrow$ & SSIM $\uparrow$ & MSE $\downarrow$
& PSNR $\uparrow$ & SSIM $\uparrow$ & MSE $\downarrow$
& PSNR $\uparrow$ & SSIM $\uparrow$ & MSE $\downarrow$
& PSNR $\uparrow$ & (M) $\downarrow$ & (G) $\downarrow$ & (s) $\downarrow$ \\
\midrule
U-shape~\cite{peng2023u} & TIP'23 & 20.54 & 0.772 & 91.77 & 24.16 & 0.846 & 81.40 & 25.48 & 0.830 & 74.20 & {23.39} & 65.60 & 66.20 & 0.063\\
GUPDM~\cite{mu2023generalized} & MM'23 & 23.99 & 0.877 & 75.49 & 27.67 & 0.867 & 56.97 & 26.21 & 0.850 & 64.45 & {25.96} & 1.49 & 95.80 & 0.696\\
Semi-UIR~\cite{huang2023contrastive} & CVPR'23 & 23.49 & 0.860 & 79.86 & 27.46 & 0.878 & 63.57 & 26.85 & 0.876 & 60.57 & {25.93} & 3.31 & 72.88 & 0.785\\
wMamba~\cite{guan2024watermamba} & arXiv'24 & 23.55 & 0.888 & 78.32 & 28.71 & 0.894 & 50.32 & 28.55 & 0.882 & 47.01 & {26.94} & 3.69 & 7.53 & 0.028\\
CECF~\cite{cong2024underwater} & AAAI'24 & 21.94 & 0.868 & 86.06 & 26.34 & 0.878 & 64.94 & 28.34 & 0.874 & 49.35 & {25.54} & 39.69 & 83.58 & 0.002\\
MambaIR~\cite{guo2024mambair} & ECCV'24 & 24.04 & 0.912 & 75.99 & 29.16 & \textcolor{blue}{0.903} & 47.97 & 29.23 & 0.892 & 43.07 & {27.48} & 25.92 & 137.57 & 0.103\\
WF-Diff~\cite{zhao2024wavelet} & CVPR'24 & 22.78 & 0.912 & 83.64 & 27.49 & 0.894 & 59.79 & 27.01 & 0.876 & 60.06 & {25.76} & 18.48 & 570.09 & 0.146\\
HCLR~\cite{zhou2024hclr} & IJCV'24 & 21.98 & 0.887 & 84.74 & 28.74 & 0.889 & 49.93 & 27.86 & 0.864 & 52.38 & {26.19} & 4.87 & 401.90 & 0.054\\
Concertormer~\cite{kuo2025efficient} & ICCV'25 & 22.55 & 0.901 & 84.27 & 26.97 & 0.893 & 60.60 & 28.53 & \textcolor{blue}{0.901} & 48.07 & {26.02} & 45.96 & 201.01 & 0.108\\
MaIR~\cite{li2025mair} & CVPR'25 & \textcolor{blue}{25.11} & \textcolor{red}{0.923} & \textcolor{green}{71.33} & \textcolor{red}{29.76} & \textcolor{red}{0.908} & \textcolor{red}{44.70} & \textcolor{blue}{29.90} & \textcolor{green}{0.899} & \textcolor{blue}{39.80} & {\textcolor{blue}{28.26}} & 7.78 & 12.59 & 0.377\\
EVSSM~\cite{kong2025efficient} & CVPR'25 & \textcolor{green}{24.93} & \textcolor{green}{0.916} & \textcolor{blue}{70.09} & \textcolor{green}{29.49} & \textcolor{green}{0.902} & \textcolor{green}{45.97} & \textcolor{green}{29.67} & 0.885 & \textcolor{green}{41.31} & {\textcolor{green}{28.03}} & 17.10 & 126.00 & 0.279\\
IA2U~\cite{11267029} & TIP'25 & 22.53 & 0.888 & 83.01 & 26.13 & 0.883 & 68.02 & 26.29 & 0.870 & 62.84 & {24.98} & 21.70 & 33.32 & 0.081\\
PyUIE~\cite{11330175} & TMM'26 & 22.94 & 0.898 & 81.10 & 26.92 & 0.882 & 61.92 & 27.07 & 0.870 & 57.61 & {25.64 }& 4.45 & 12.92 & 0.039\\
DiffColor~\cite{11353942} & TMM'26 & 23.36 & 0.902 & 79.11 & 28.07 & 0.891 & 53.99 & 28.58 & 0.884 & 46.25 & {26.67 }& 22.16 & 87.95 & 0.157\\
DTIUIE~\cite{11433506} & TIP'26 & 24.57 & 0.911 & 71.39 & 28.18 & 0.893 & 53.89 & 28.10 & 0.880 & 49.64 & {26.95 } & 50.81 & 385.93 & 0.112\\
CLUIE (Ours) & - & \textcolor{red}{25.53} & \textcolor{blue}{0.921} & \textcolor{red}{67.87} & \textcolor{blue}{29.60} & \textcolor{green}{0.902} & \textcolor{blue}{45.16} & \textcolor{red}{30.74} & \textcolor{red}{0.903} & \textcolor{red}{35.09} & {\textcolor{red}{28.62}} & 4.39 & 14.26 & 0.115\\
\bottomrule
\end{tabular}
\end{table*}
Following the branch-wise mixing strategy in visual RWKV, the original token feature $\mathbf{X}$ and the local feature $\mathbf{X}^{loc}$ are mixed separately for Key, Value, and Receptance:
\begin{equation}
\mathbf{X}^{(b)}
=
\mathbf{X}
\odot
\mathbf{m}_{b}
+
\mathbf{X}^{loc}
\odot
\left(
1-\mathbf{m}_{b}
\right),
\quad
b\in\{k,v,r\},
\label{eq:dmlp_branch_mixing}
\end{equation}
where $m_b\in\mathbb{R}^{1\times1\times C}$ is a learnable channel-wise mixing parameter. The branch-specific features are then projected as
\begin{equation}
\mathbf{k}=\mathbf{X}^{(k)}
\mathbf{W}_{k},
\mathbf{v}=\mathbf{X}^{(v)}\mathbf{W}_{v},
\mathbf{r}
=\mathbf{X}^{(r)}\mathbf{W}_{r},
\mathbf{s}
=\sigma\left(\mathbf{r}\right),
\label{eq:dmlp_kvr_projection}
\end{equation}
where $\mathbf{s}$ is the receptance gate. DMLP therefore changes the local information supplied to the RWKV branches, while keeping the WKV aggregation form unchanged.

\section{Experiments}
\label{sec:experiments}

\subsection{Experimental Setup}
\subsubsection{Datasets and Evaluation Metrics}
We evaluate CLUIE on paired and unpaired underwater image enhancement benchmarks. For training, we use a mixed paired dataset consisting of 800 image pairs from UIEB and 3,879 image pairs from LSUI, resulting in 4,679 training pairs in total \cite{li2019underwater,peng2023u}. For full-reference evaluation, the trained model is tested on 90 images from UIEB, 400 images from LSUI, and 481 images from EUVP \cite{li2019underwater,peng2023u}. The training and testing images are kept disjoint.

For full-reference evaluation, we report PSNR, SSIM, and MSE to measure pixel fidelity and structural consistency \cite{wang2004image}. For unpaired real-world evaluation, UCIQE and UIQM are commonly adopted in underwater image quality assessment \cite{yang2015underwater,panetta2015human}. We also report MUSIQ and NIMA to provide learning-based perceptual quality assessment \cite{9710973,8352823}.


\subsubsection{Implementation Details}
CLUIE is implemented in PyTorch and trained on two NVIDIA RTX 3080 GPUs with distributed data parallelism. The base channel number is set to 24 and the batch size is 6. The model is trained for 1000 epochs. The initial learning rate is $2\times10^{-4}$ and the minimum learning rate is $1\times10^{-6}$. Gradient clipping is applied with a threshold of $0.001$, and the random seed is fixed to 42.

Unless otherwise specified, the number of semantic groups in CSDR is set to $K=8$. In DMLP, the pseudo-dark response window is fixed to $3\times3$, while the depth-wise local propagation windows are set to $3\times3$, $5\times5$, and $7\times7$ according to layer depth.

\subsection{Quantitative and Qualitative Comparison}

\subsubsection{Full-reference Comparison}
We compare CLUIE with representative UIE and image restoration methods, including CNN-based, Transformer-based, diffusion-based, state-space-based, and recent task-specific approaches \cite{11330175,peng2023u,11267029,mu2023generalized,huang2023contrastive,cong2024underwater,zhao2024wavelet,guan2024watermamba,zhou2024hclr,zhou2025devil,kuo2025efficient,li2025mair,kong2025efficient,11433506}. Table~\ref{tab:full_ref} reports the full-reference comparison on UIEB, LSUI, and EUVP. CLUIE achieves the best PSNR and MSE on UIEB and EUVP, and obtains competitive results on LSUI. Specifically, on UIEB, CLUIE reaches 25.53 dB PSNR and 67.87 MSE, outperforming the compared methods in pixel-level fidelity. On EUVP, it obtains 30.74 dB PSNR and 35.09 MSE, indicating that the model also generalizes to another paired benchmark with different degradation characteristics. On LSUI, MaIR and EVSSM achieve slightly higher PSNR or MSE, while CLUIE remains among the top-performing methods with a smaller parameter size than many Transformer- or diffusion-based models.

These results are consistent with the motivation of the proposed recurrent trajectory modeling. UIEB and EUVP contain diverse color casts, veiling effects, and local contrast variations, where restoration often requires interactions among spatially separated but degradation-related regions. Compared with fixed spatial or spectral traversal, CSDR reorganizes the WKV propagation path according to feature-level relations, which helps the recurrent state aggregate information from content-related regions. Meanwhile, DMLP supplies local structural responses before token reordering, reducing the risk of losing edge and texture continuity. The gains in PSNR and MSE therefore suggest that the combination of content-adaptive long-range propagation and local structural compensation improves restoration fidelity.

\begin{figure*}[!htb]
\centering
\includegraphics[width=1.0\textwidth]{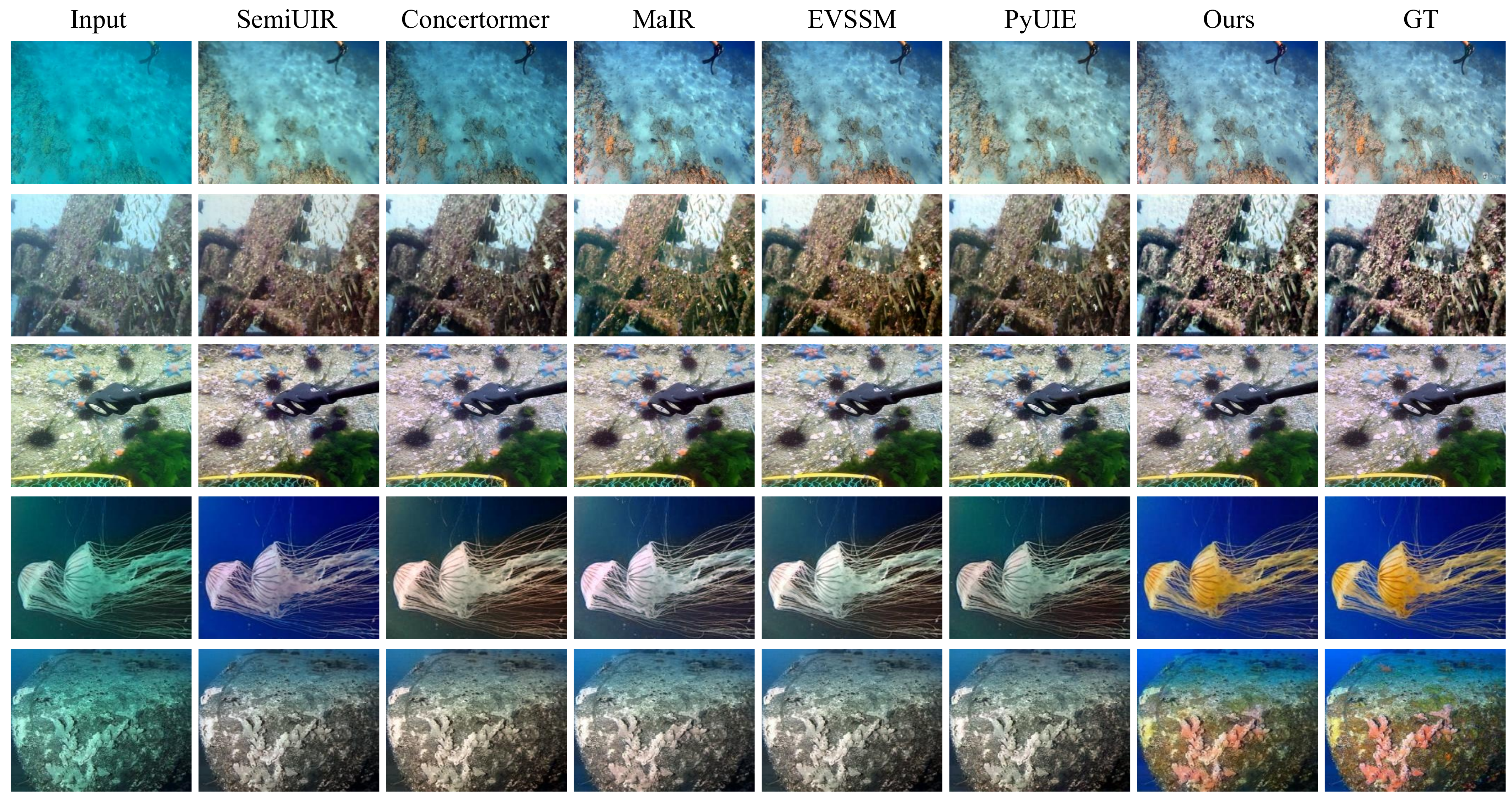}\vspace{-2mm}
\caption{Full-reference visual comparison on UIEB, LSUI, and EUVP. 
CLUIE produces color-consistent results with relatively clear local structures under different underwater degradation conditions.}\vspace{-2mm}
\label{fig:qualitative_full_ref}
\end{figure*}

It is also worth noting that CLUIE does not rely on a large model size. With 4.39M parameters and 14.26G FLOPs, it is more compact than several recent Transformer- and diffusion-based baselines. Although its inference time is not the lowest among all methods, the accuracy-complexity trade-off is favorable, especially when compared with methods that require substantially larger FLOPs or parameters. This supports the use of RWKV-based recurrent modeling as an efficient alternative for high-resolution underwater restoration.

Fig.~\ref{fig:qualitative_full_ref} presents visual comparisons on paired benchmarks. Several competing methods can improve global brightness or contrast but may leave residual green/blue color casts, over-correct foreground regions, or smooth fine textures. CLUIE produces more balanced color correction in the selected examples and preserves clearer object boundaries and local details. This behavior is related to the two components of the CRWKV block. CSDR allows recurrent aggregation to follow feature-related regions rather than a fixed raster path, which benefits global color consistency across non-adjacent regions. DMLP injects modulated local responses before reordering, which helps maintain texture and boundary continuity after long-range recurrent aggregation. The visual results are therefore consistent with the quantitative improvements in Table~\ref{tab:full_ref}.



\begin{table*}[!htb]
\tabcolsep 0.032 in
\centering
\caption{No-reference comparison with representative underwater image enhancement methods and image restoration baselines. Top-1/2/3 values are marked in \textcolor{red}{red}, \textcolor{blue}{blue}, and \textcolor{green}{green}.}
\label{tab:no_ref}
\renewcommand{\arraystretch}{1.05}
\begin{tabular}{lc*{12}{c}}
\toprule
\multirow{2}{*}{\raisebox{-0.6ex}{Method}} & \multirow{2}{*}{\raisebox{-0.6ex}{Source}}
& \multicolumn{4}{c}{C60}
& \multicolumn{4}{c}{Seathru}
& \multicolumn{4}{c}{UCCS} \\
\cmidrule(lr){3-6} \cmidrule(lr){7-10} \cmidrule(lr){11-14}
& & UCIQE $\uparrow$ & UIQM $\uparrow$ & MUSIQ $\uparrow$ & NIMA $\uparrow$
& UCIQE $\uparrow$ & UIQM $\uparrow$ & MUSIQ $\uparrow$ & NIMA $\uparrow$
& UCIQE $\uparrow$ & UIQM $\uparrow$ & MUSIQ $\uparrow$ & NIMA $\uparrow$ \\
\midrule
U-shape~\cite{peng2023u} & TIP'23
& 0.545 & 2.660 & 44.446 & 4.011
& 0.466 & 2.835 & 25.445 & 3.386
& 0.547 & \textcolor{green}{3.044} & 29.889 & 3.421 \\
GUPDM~\cite{mu2023generalized} & MM'23
& 0.550 & 2.602 & 43.440 & 3.886
& 0.459 & 2.800 & 39.989 & 3.711
& 0.552 & 2.991 & 30.996 & 3.416 \\
Semi-UIR~\cite{huang2023contrastive} & CVPR'23
& \textcolor{green}{0.585} & 2.648 & \textcolor{red}{48.140} & 4.304
& 0.507 & 2.919 & \textcolor{red}{48.799} & 4.263
& 0.571 & 2.951 & \textcolor{red}{33.233} & 3.485 \\
wMamba~\cite{guan2024watermamba} & arXiv'24
& 0.580 & \textcolor{red}{2.775} & 45.975 & 3.999
& 0.503 & \textcolor{green}{3.010} & 38.402 & 3.581
& 0.569 & \textcolor{green}{3.044} & 32.705 & 3.481 \\
CECF~\cite{cong2024underwater} & AAAI'24
& 0.583 & 2.695 & 43.994 & 3.923
& \textcolor{red}{0.536} & 2.967 & 41.323 & 3.770
& 0.568 & 2.948 & 30.957 & 3.441 \\
WF-Diff~\cite{zhao2024wavelet} & CVPR'24
& 0.584 & \textcolor{green}{2.715} & 47.693 & \textcolor{blue}{4.346}
& 0.501 & \textcolor{blue}{3.028} & 44.818 & 4.391
& \textcolor{red}{0.576} & 3.041 & 31.944 & \textcolor{blue}{3.610} \\
HCLR~\cite{zhou2024hclr} & IJCV'24
& 0.566 & \textcolor{blue}{2.737} & 46.050 & 4.138
& 0.466 & 2.141 & 36.310 & 3.861
& 0.565 & 2.996 & 30.725 & 3.422 \\
Concertormer~\cite{kuo2025efficient} & ICCV'25
& 0.563 & 2.632 & 47.573 & 4.231
& 0.427 & 2.057 & 35.062 & 3.719
& 0.550 & 2.957 & \textcolor{green}{32.757} & 3.467 \\
MaIR~\cite{li2025mair} & CVPR'25
& \textcolor{red}{0.589} & 2.564 & \textcolor{green}{47.717} & \textcolor{red}{4.354}
& \textcolor{blue}{0.528} & 2.217 & 41.187 & 4.191
& 0.555 & 2.926 & 30.890 & \textcolor{red}{3.657} \\
EVSSM~\cite{kong2025efficient} & CVPR'25
& 0.584 & 2.640 & 47.087 & 4.302
& 0.485 & 2.823 & \textcolor{blue}{46.576} & \textcolor{green}{4.403}
& \textcolor{green}{0.572} & 2.963 & 32.003 & \textcolor{green}{3.571} \\
IA2U~\cite{11267029} & TIP'25
& 0.577 & 2.689 & 47.289 & 4.129
& \textcolor{green}{0.519} & 2.764 & 43.815 & \textcolor{blue}{4.458}
& 0.564 & 3.005 & 32.589 & 3.424 \\
PyUIE~\cite{11330175} & TMM'26
& 0.575 & 2.543 & 45.529 & 4.054
& 0.467 & 2.324 & 31.332 & 3.963
& 0.562 & 2.959 & 30.788 & 3.413 \\
DiffColor~\cite{11353942 }& TMM'26
& 0.574 & 2.590 & 46.575 & 4.245
& 0.471 & 2.850 & 36.990 & 4.246
& 0.569 & 2.955 & 31.080 & 3.553 \\
DTIUIE~\cite{11433506} & TIP'26
& 0.572 & 2.651 & 47.153 & 4.127
& 0.496 & 2.892 & 40.881 & 3.901
& 0.564 & \textcolor{blue}{3.071} & 32.554 & 3.462 \\
CLUIE (Ours) & -
& \textcolor{blue}{0.586} & 2.607 & \textcolor{blue}{47.719} & \textcolor{green}{4.331}
& \textcolor{green}{0.519} & \textcolor{red}{3.071} & \textcolor{green}{45.246} & \textcolor{red}{4.506}
& \textcolor{blue}{0.574} & \textcolor{red}{3.082} & \textcolor{blue}{33.117} & 3.507 \\
\bottomrule
\end{tabular}\vspace{-2mm}
\end{table*}

\begin{figure*}[!htb]
\centering
\includegraphics[width=\textwidth]{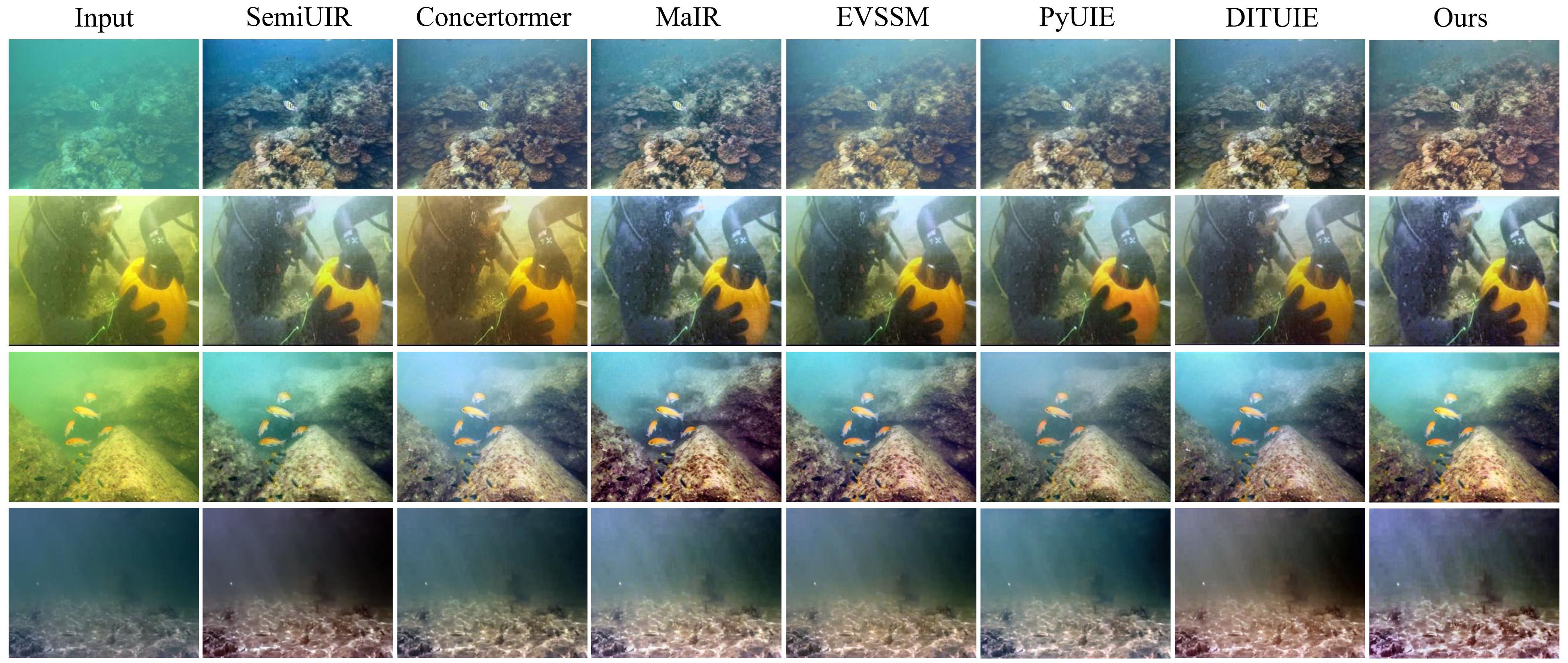}\vspace{-2mm}
\caption{No-reference visual comparison on real-world underwater images. 
CLUIE provides balanced enhancement under different illumination and color-degradation conditions.}\vspace{-4mm}
\label{fig:no_ref_visual}
\end{figure*}

\subsubsection{No-reference Evaluation}
We further evaluate different methods on real-world underwater datasets without paired references. Table~\ref{tab:no_ref} reports UCIQE, UIQM, MUSIQ, and NIMA on C60, Seathru, and UCCS. Since no-reference metrics emphasize different perceptual factors, a single metric may not fully represent visual quality. UCIQE is more related to chroma, saturation, and contrast, UIQM combines colorfulness, sharpness, and contrast, while MUSIQ and NIMA provide learning-based perceptual or aesthetic assessment. Therefore, we analyze the overall tendency across datasets and metrics. 

CLUIE shows competitive no-reference performance on all three datasets. On Seathru, it obtains the best UIQM and NIMA, and ranks among the top methods in UCIQE and MUSIQ. On UCCS, it achieves the best UIQM and the second-best UCIQE and MUSIQ. On C60, it ranks near the top in UCIQE, MUSIQ, and NIMA, although its UIQM is lower than several baselines. These results indicate that CLUIE does not simply optimize one specific perceptual criterion. Although CLUIE does not consistently rank first on every no-reference metric, its results remain stable across different datasets and perceptual criteria.



Fig.~\ref{fig:no_ref_visual} shows representative results on real-world underwater images. Some methods introduce excessive saturation or incomplete color correction, especially in images with uneven illumination. Others improve visibility but suppress local textures or produce unnatural color transitions. CLUIE provides more balanced enhancement in the shown examples, with improved foreground visibility and fewer obvious color artifacts. This observation is consistent with the no-reference metrics in Table~\ref{tab:no_ref}. The results suggest that the proposed trajectory reordering does not overfit to paired training references, and the local propagation module helps preserve visible structures in unpaired real-world scenes.

\begin{figure}[!t]
\centering
\includegraphics[width=\columnwidth]{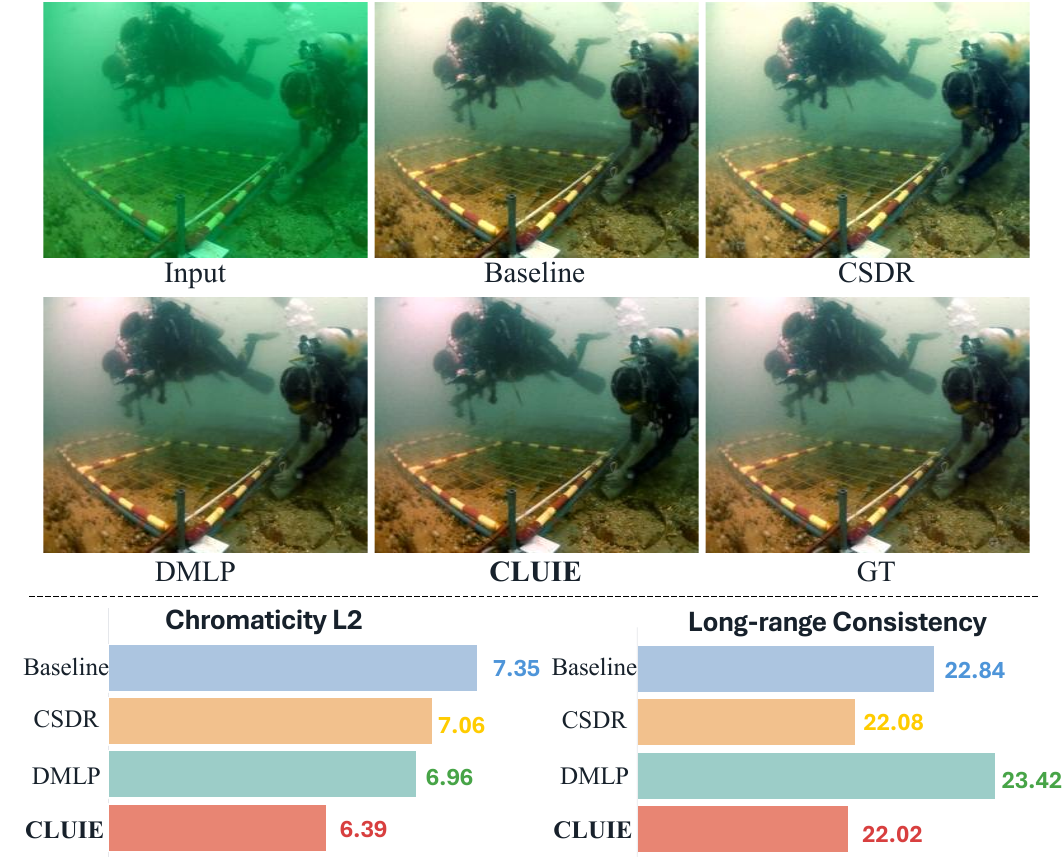}\vspace{-2mm}
\caption{Visual and statistical comparison of CSDR and DMLP on UIEB using color-composition and long-range consistency metrics. Lower values indicate better restoration.}
\label{fig:csdr_dmlp_vis}
\end{figure}

\subsection{Ablation Study}
To evaluate the contribution of each proposed component, we conduct ablation studies on UIEB under the same training settings as the full model. All variants are trained under the same setting as the full model unless otherwise specified. PSNR, SSIM, and MSE are reported.

\subsubsection{Effectiveness of CSDR and DMLP}
We first evaluate the contributions of CSDR and DMLP. The baseline removes both modules, uses raster-order RWKV propagation, and adopts the original Q-shift operation for local interaction \cite{duan2025vision,du2024exploring}. CSDR and DMLP are then added separately and jointly.


\begin{table}[!t]
\tabcolsep 0.02 in
\centering
\caption{Ablation study on the effectiveness of CSDR and DMLP on UIEB.}
\label{tab:ablation_components}
\resizebox{0.99\linewidth}{!}{
\begin{tabular}{lcccccc}
\toprule
Variant & PSNR~$\uparrow$ & SSIM~$\uparrow$ & MSE~$\downarrow$& Par.(M)~$\downarrow$ & FLOPs(G)~$\downarrow$ & Time(s)~$\downarrow$ \\
\midrule
Baseline & 24.23 & 0.914 & 77.62 & 4.29 & 13.12 & 0.084 \\
Baseline + CSDR & 24.53 & 0.917 & 74.93 & 4.29 & 13.61 & 0.107 \\
Baseline + DMLP & 24.85 & 0.912 & 71.16 & 4.39 & 13.81 & 0.085 \\
\textbf{CLUIE (Ours)} & \textbf{25.53} & \textbf{0.921} & \textbf{67.87} & \textbf{4.39} & \textbf{14.26} & \textbf{0.115} \\
\bottomrule
\end{tabular}
}
\end{table}

As reported in Table~\ref{tab:ablation_components}, CSDR improves the PSNR from 24.23 dB to 24.53 dB and reduces the MSE from 77.62 to 74.93, indicating that replacing fixed raster-order propagation with content-adaptive reordering enables more effective recurrent state aggregation. Since CSDR mainly performs content-adaptive clustering and feature reordering, it introduces no additional learnable parameters. Its computational overhead comes primarily from indexing and grouping operations, which moderately increase FLOPs and inference time. DMLP alone further increases the PSNR to 24.85 dB and decreases the MSE to 71.16, showing the effectiveness of modulated local propagation. Unlike CSDR, DMLP adds a small number of learnable parameters, increasing the model size from 4.29M to 4.39M. However, its computation is dominated by dense, hardware-friendly operations, so its inference time remains close to the baseline. Notably, the SSIM of DMLP alone is slightly lower than that of the baseline, suggesting that local propagation cannot fully maintain structural consistency when the recurrent trajectory remains fixed. Combining CSDR and DMLP yields the best results across all metrics. Although the full CLUIE has the highest complexity among the variants, it adds only 0.10M parameters over the baseline and achieves consistent improvements in PSNR, SSIM, and MSE. These results show that CSDR and DMLP provide complementary gains: CSDR adapts long-range recurrent propagation according to content similarity, whereas DMLP reinforces local spatial responses before recurrent aggregation.

Fig.~\ref{fig:csdr_dmlp_vis} further explains this complementarity using two diagnostic measures: Chromaticity L2, computed from brightness-normalized RGB ratios, and long-range distance, defined as the residual distance between spatially distant pixels assigned to the same CSDR cluster. DMLP reduces Chromaticity L2, indicating improved local color restoration, but increases the long-range distance, which suggests that local modulation alone may weaken consistency across distant regions. In contrast, CSDR reduces the long-range distance by promoting content-adaptive propagation among correlated regions. Their combination achieves the lowest values on both measures, confirming that DMLP mainly improves local color composition while CSDR strengthens global structural consistency.

\subsubsection{Analysis of CSDR}
We analyze CSDR from three aspects: trajectory construction, the number of semantic groups, and feature-response alignment.

First, we compare different trajectory construction strategies. The raster-order variant keeps the original spatial scan. The variant without intra-cluster traversal applies cluster-level ordering but does not preserve spatial continuity inside each cluster. The CLUIE further uses connected-component-based intra-cluster traversal.



\begin{table}[!t]
\tabcolsep 0.2 in
\centering
\caption{Ablation study on recurrent trajectory reconstruction strategies in CSDR on UIEB.}
\label{tab:ablation_csdr}
\setlength{\tabcolsep}{4.2pt}
\renewcommand{\arraystretch}{1.05}
\begin{tabular}{lccc}
\toprule
Trajectory & PSNR $\uparrow$ & SSIM $\uparrow$ & MSE $\downarrow$ \\
\midrule
Raster order & 24.85 & 0.912 & 71.16 \\
CSDR w/o intra-cluster & 25.23 & 0.917 & 68.78 \\
CSDR w random-sorting & 25.28 & 0.916 & 67.93 \\
\textbf{CLUIE (Ours)} & \textbf{25.53} & \textbf{0.921} & \textbf{67.87} \\
\bottomrule
\end{tabular}
\end{table}

Table~\ref{tab:ablation_csdr} evaluates the effect of recurrent trajectory construction. Replacing raster-order propagation with cluster-level ordering improves the PSNR from 24.85 dB to 25.23 dB and the SSIM from 0.912 to 0.917, confirming that the token order in WKV state accumulation has a direct impact on cross-region information propagation. Compared with a fixed raster scan, cluster-level ordering allows recurrent states to aggregate restoration cues along more semantically related regions. Further introducing spatially coherent intra-cluster traversal raises the performance to 25.53 dB PSNR and 0.921 SSIM, indicating that content-adaptive reordering should preserve local continuity within each cluster. Thus, an effective recurrent trajectory should combine global feature-level grouping with local spatial coherence. The random sorting among clusters further validates the necessity of relation-guided inter-cluster ordering. Replacing the relation matrix with random cluster transitions decreases the PSNR from 25.53 dB to 25.28 dB and the SSIM from 0.921 to 0.916, while increasing the MSE from 67.87 to 67.93. This degradation shows that the relation matrix provides meaningful inter-cluster ordering cues, enabling recurrent states to propagate along coherent semantic trajectories rather than arbitrary cluster transitions.

Table~\ref{tab:ablation_k} analyzes the grouping granularity. The best performance is obtained when $K=8$. A smaller $K$ may merge regions with different degradation responses into the same group, reducing the flexibility of the recurrent trajectory. A larger $K$ may over-fragment coherent regions and introduce unstable transitions. The performance trend suggests that a moderate number of semantic groups provides a better balance between content discrimination and trajectory stability.



\begin{table}[!t]
\tabcolsep 0.15 in
\centering
\caption{Effect of the number of semantic groups $K$ in CSDR on UIEB.}
\label{tab:ablation_k}
\renewcommand{\arraystretch}{1.05}
\begin{tabular}{cccc}
\toprule
$K$ & PSNR $\uparrow$ & SSIM $\uparrow$ & MSE $\downarrow$ \\
\midrule
6 & 25.22 & 0.913 & 68.54 \\
\textbf{8} & \textbf{25.53} & \textbf{0.921} & \textbf{67.87} \\
10 & 25.18 & 0.917 & 69.58 \\
\bottomrule
\end{tabular}
\end{table}


\begin{figure}[!t]
\centering
\includegraphics[width=\columnwidth]{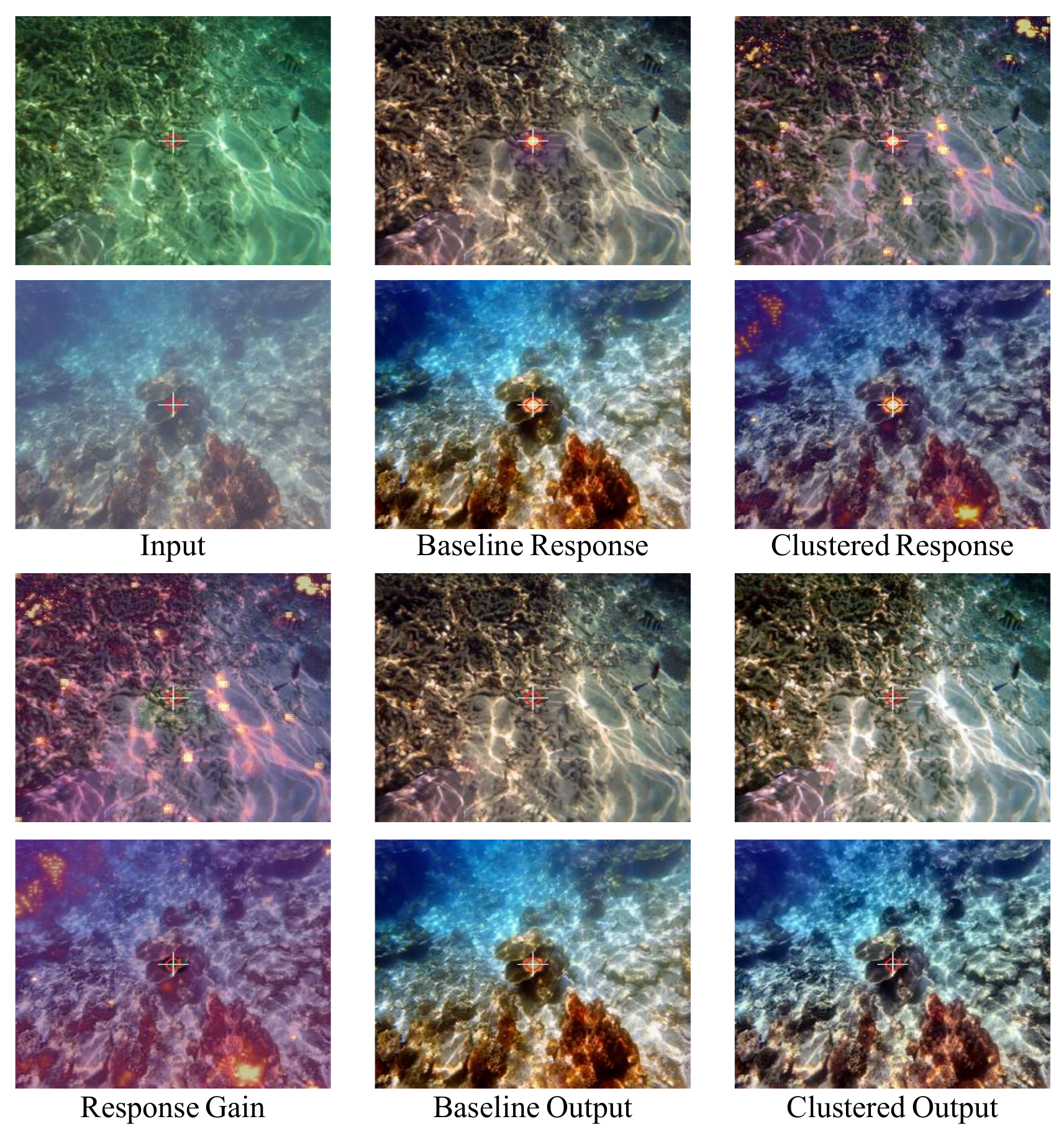}\vspace{-2mm}
\caption{Response visualization of CSDR. Compared with raster-order propagation, clustered state propagation produces more coherent activations over feature-related regions.}
\label{fig:csdr_response_visualization}
\end{figure}

\begin{figure}[!t]
\centering
\includegraphics[width=\columnwidth]{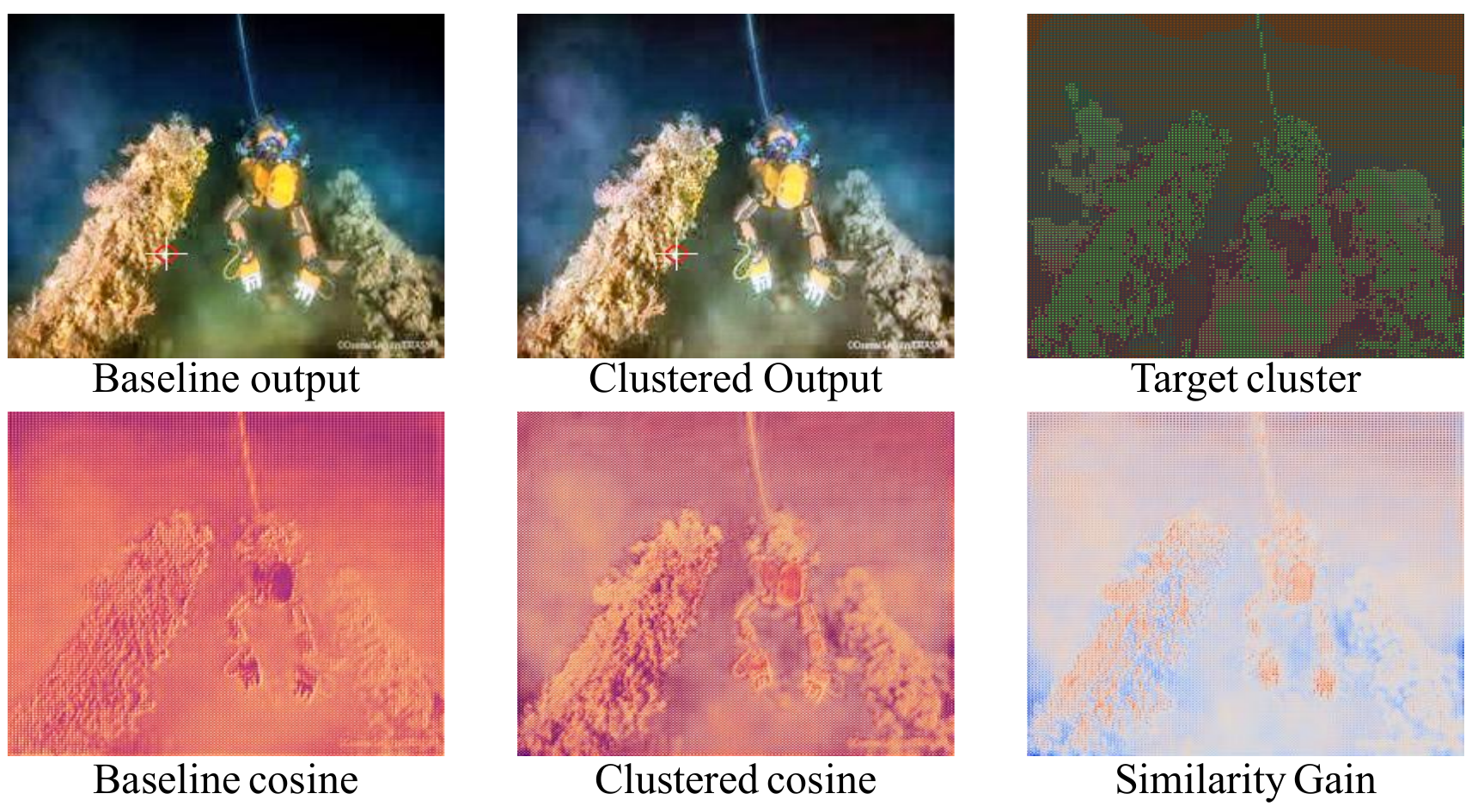}\vspace{-2mm}
\caption{Feature-level visualization of clustering effectiveness. The target cluster contains tokens assigned to the same semantic group as the queried patch. CSDR yields stronger responses over feature-related regions.}
\label{fig:feature_cosine_cluster}
\end{figure}

\begin{figure}[!t]
\centering
\includegraphics[width=\columnwidth]{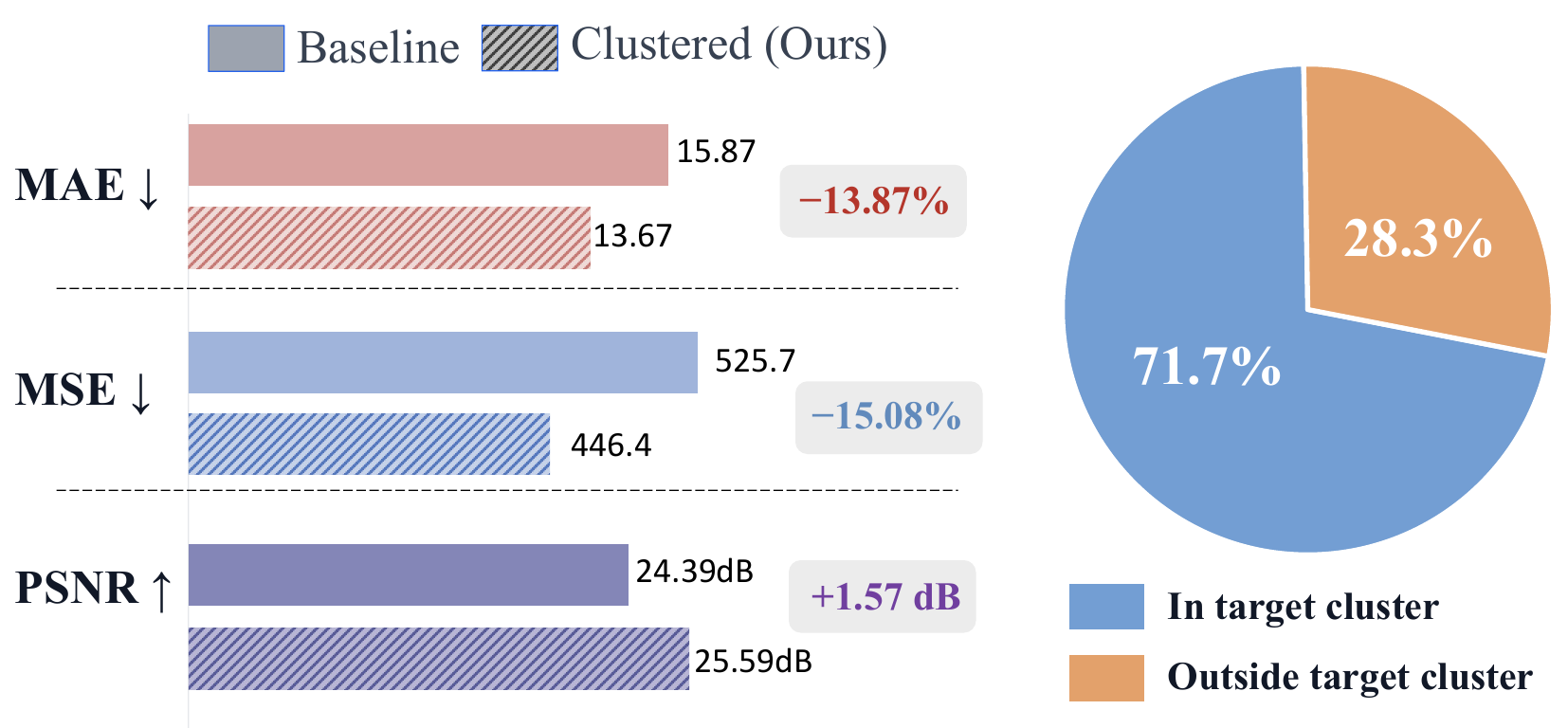}
\caption{Statistical analysis of feature-response alignment. Left: restoration quality on high-response regions. Right: spatial distribution of selected high-response patches with respect to the target cluster.}
\label{fig:feature_cluster_statistics}
\end{figure}

The visual and statistical analyses in Figs.~\ref{fig:csdr_response_visualization}--\ref{fig:feature_cluster_statistics} further explain why CSDR improves restoration. Compared with raster-order propagation, clustered propagation produces more coherent responses over feature-related regions. For the top-20\% high-response regions, CSDR reduces MAE from 15.875 to 13.672 and MSE from 525.7 to 446.4, while increasing PSNR from 24.39 dB to 25.59 dB. In addition, 71.7\% of high-response patches fall inside the corresponding target cluster. These results indicate that the constructed trajectory is aligned with feature-level correlations rather than being an arbitrary permutation. This supports the central claim that CSDR makes RWKV state propagation more consistent with region-dependent restoration relations in underwater images.

Fig.~\ref{fig:feature_cosine_cluster} shows that spatially separated tokens within the same cluster can have consistent feature responses. After clustered propagation, the cosine response becomes more concentrated in the target cluster, indicating that CSDR promotes interactions among feature-related regions.

\subsubsection{Analysis of DMLP}

We further analyze DMLP. The Q-shift variant uses the original fixed shift operation in visual RWKV. The local-structural-only variant removes the pseudo-dark modulation and directly injects depth-wise local responses. The pseudo-dark-only variant removes the learnable local propagation branch and uses only the pseudo-dark response for modulation. The CLUIE combines both components.


\begin{table}[!t]
\tabcolsep 0.2 in
\centering
\caption{Ablation study on the component design of DMLP on UIEB.}
\label{tab:ablation_dmlp}
\setlength{\tabcolsep}{4.2pt}
\renewcommand{\arraystretch}{1.05}
\begin{tabular}{lccc}
\toprule
Design & PSNR $\uparrow$ & SSIM $\uparrow$ & MSE $\downarrow$ \\
\midrule
Q-shift & 24.05 & 0.903 & 75.32 \\
Local structural only & 24.95 & 0.914 & 70.83 \\
Pseudo-dark only & 24.94 & 0.916 & 70.83 \\
\textbf{CLUIE (Ours)} & \textbf{25.53} & \textbf{0.921} & \textbf{67.87} \\
\bottomrule
\end{tabular}
\end{table}

Table~\ref{tab:ablation_dmlp} evaluates the design of DMLP. Replacing Q-shift with learnable local structural propagation improves PSNR from 24.05 dB to 24.95 dB and SSIM from 0.903 to 0.914. This shows that fixed channel-group shifting is limited for modeling local structures in UIE, where edges, textures, and color transitions vary spatially. The pseudo-dark-only variant also improves the baseline, suggesting that the feature-space pseudo-dark response provides useful spatial modulation for local feature adjustment. The CLUIE achieves the best performance, with 25.53 dB PSNR and 0.921 SSIM. The gain over either single component indicates that learnable local aggregation and pseudo-dark modulation serve different roles: the former extracts local structural cues, while the latter controls their spatial injection strength.

\begin{table}[!t]
\tabcolsep 0.2 in
\centering
\caption{Ablation study on the local propagation window in the local structural propagation branch on UIEB.}
\label{tab:ablation_window}
\setlength{\tabcolsep}{4.2pt}
\renewcommand{\arraystretch}{1.05}
\begin{tabular}{lccc}
\toprule
Window & PSNR $\uparrow$ & SSIM $\uparrow$ & MSE $\downarrow$ \\
\midrule
Fixed $3\times3$ & 24.61 & 0.912 & 72.94 \\
Fixed $5\times5$ & 24.68 & 0.912 & 72.41 \\
Layer--based $3/5/7$ & \textbf{24.95} & \textbf{0.914} & \textbf{70.83} \\
\bottomrule
\end{tabular}
\end{table}


Table~\ref{tab:ablation_window} further shows that the layer-based $3/5/7$ window outperforms fixed $3\times3$ and $5\times5$ windows when only the local structural branch is used. This result is consistent with the hierarchical nature of encoder-decoder restoration networks. Shallow layers require compact neighborhoods to preserve fine textures, while deeper layers benefit from broader local context for regional structure recovery. A fixed window applies the same locality assumption to all stages, whereas the layer-based setting provides a simple scale adaptation.

\begin{figure}[!t]
\centering
\includegraphics[width=\columnwidth]{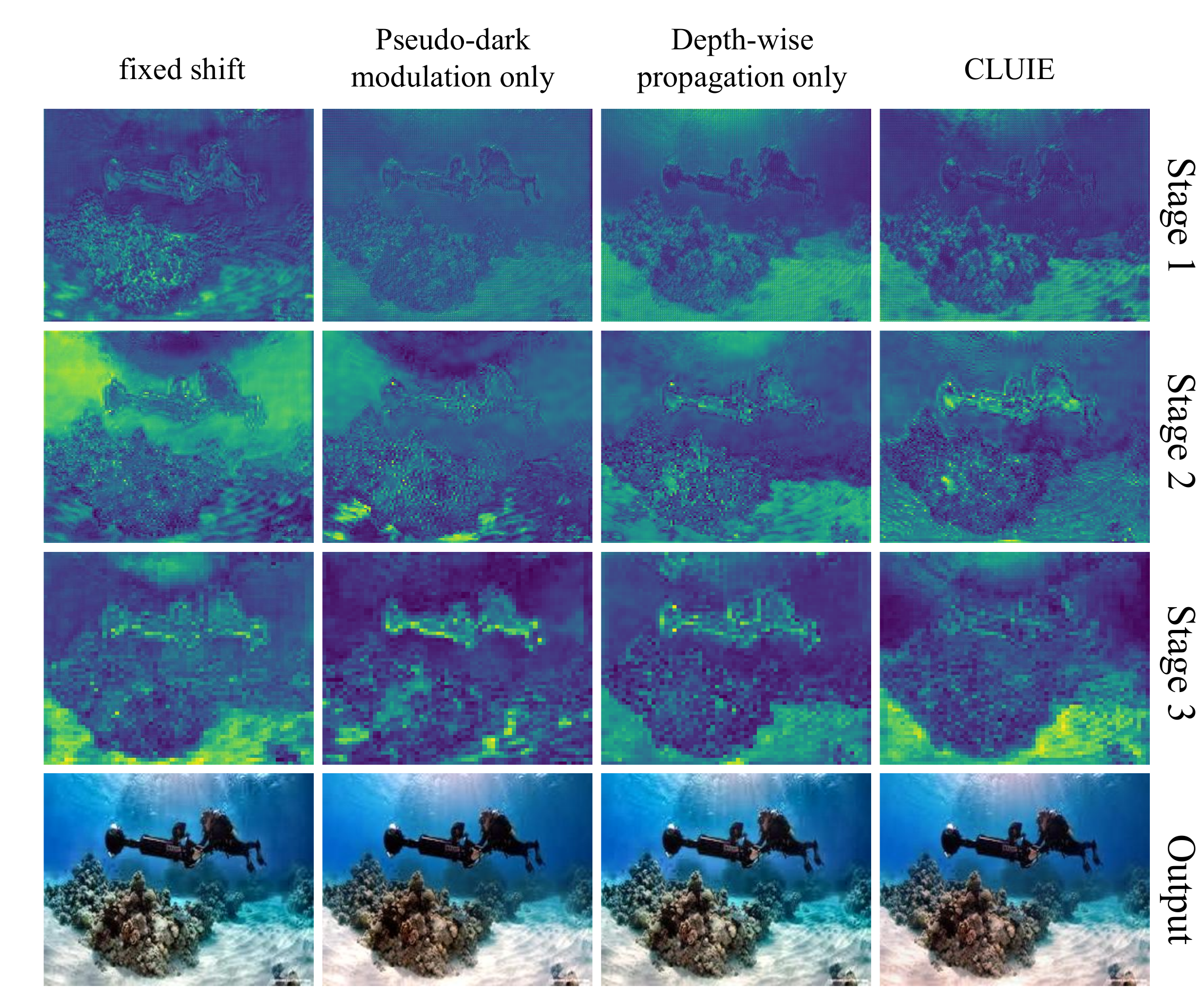}\vspace{-2mm}
\caption{Feature visualization of different DMLP designs. Our CLUIE produces more continuous local responses across layers than fixed shift.}
\label{fig:dmlp_feature_map}
\end{figure}

Fig.~\ref{fig:dmlp_feature_map} visualizes feature responses from different local propagation designs. Compared with fixed shift, DMLP produces more continuous activations around object boundaries and textured regions. This indicates that learnable local propagation can better preserve spatial organization than fixed channel relocation.


\begin{figure}[!t]
\centering
\includegraphics[width=\columnwidth]{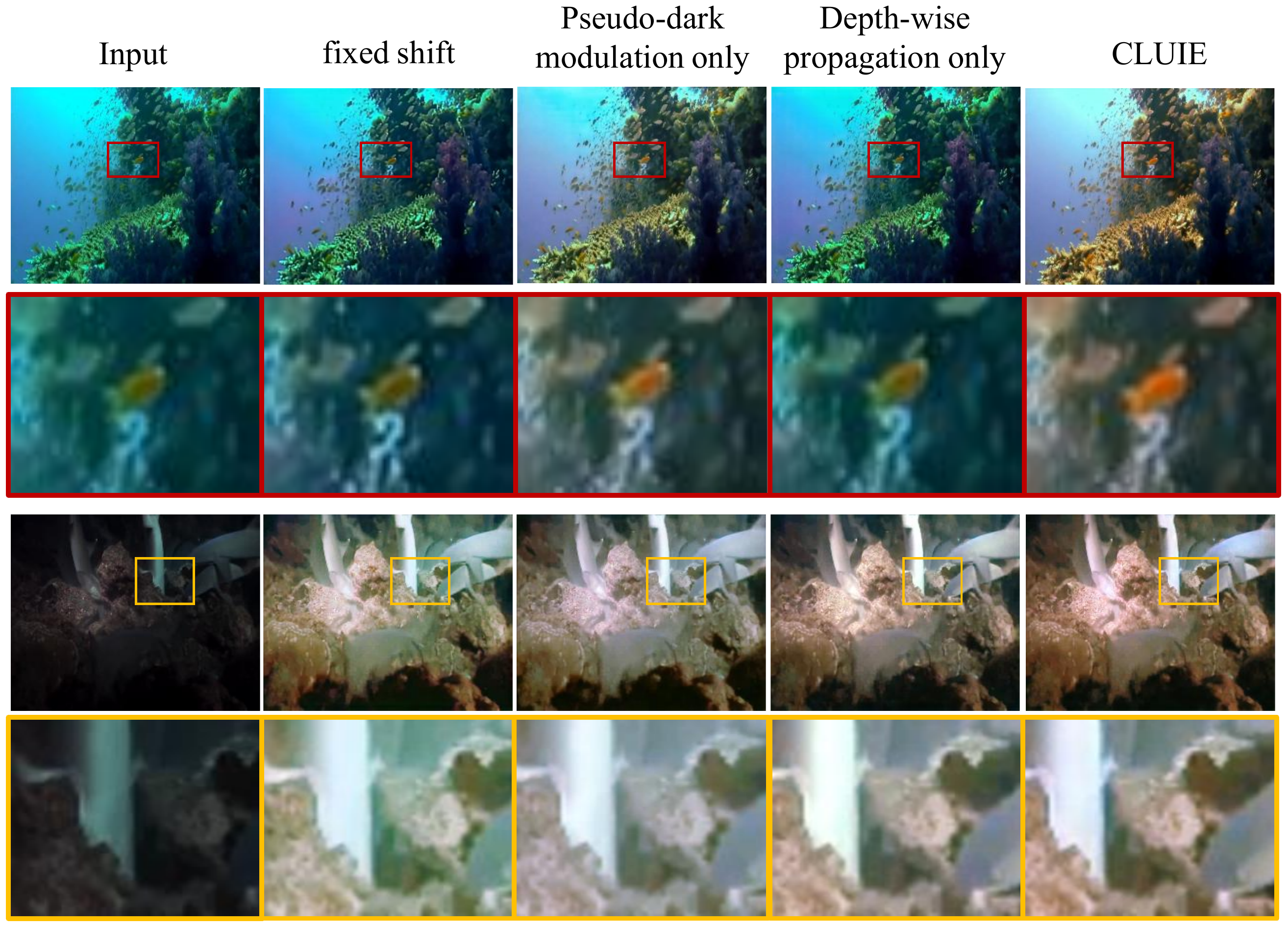}\vspace{-2mm}
\caption{Qualitative ablation of DMLP designs. Combining pseudo-dark modulation with local structural propagation gives clearer local structures and more balanced color restoration than either component alone.}
\label{fig:dmlp_visual_ablation}
\end{figure}

Fig.~\ref{fig:dmlp_visual_ablation} compares restored images from different DMLP variants. Pseudo-dark modulation alone improves spatial adaptation but lacks learnable structural aggregation. Local structural propagation alone enhances details but does not explicitly modulate spatial injection strength. By contrast, the complete DMLP better preserves foreground structures and local color consistency.

\begin{figure}[!t]
\centering
\includegraphics[width=\columnwidth]{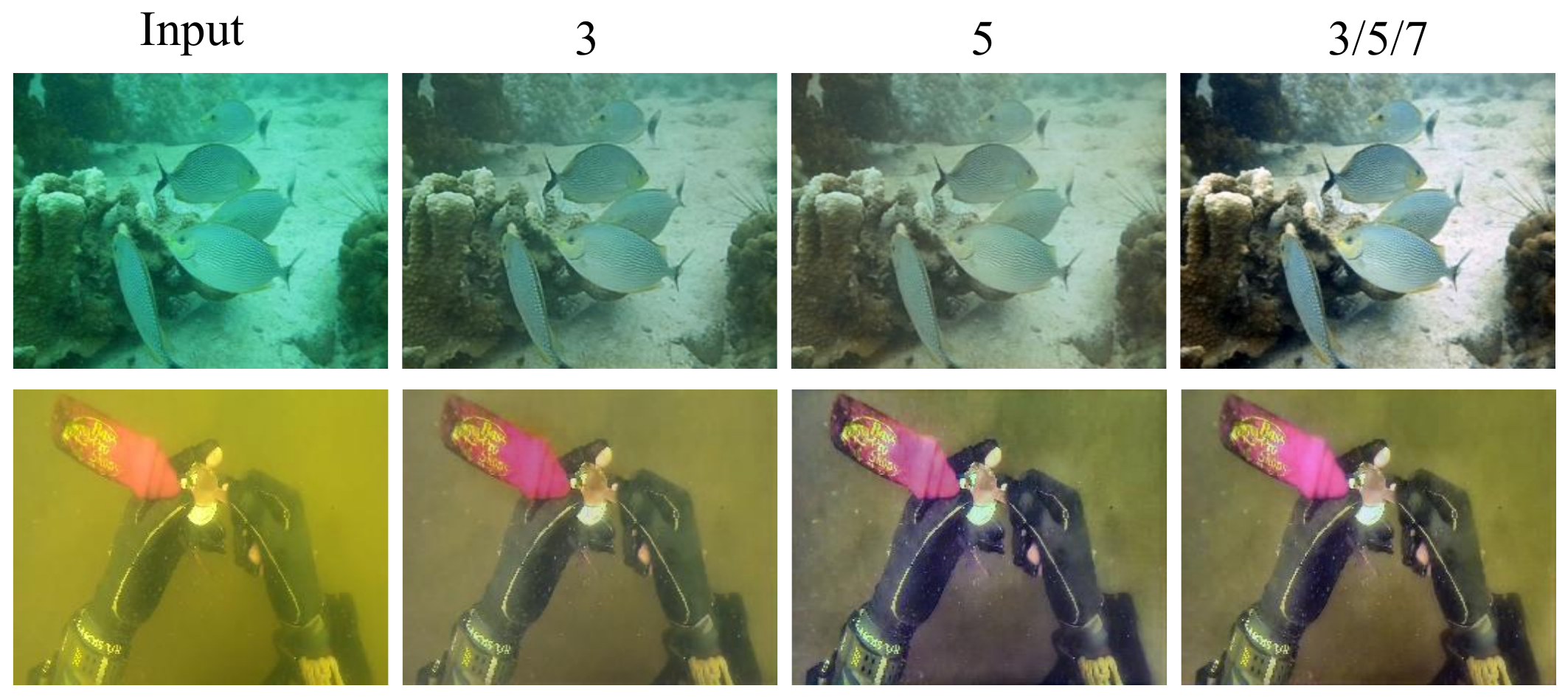}\vspace{-2mm}
\caption{Visualization of different local propagation window settings. The layer-based $3/5/7$ (CLUIE) strategy balances fine-detail preservation and regional structural propagation.}
\label{fig:shift_window_visualization}
\end{figure}

Fig.~\ref{fig:shift_window_visualization} compares visual results from different propagation windows. The fixed $3\times3$ setting focuses on local details but suffers from insufficient regional consistency, while the fixed $5\times5$ setting increases the neighborhood range but ignores the heterogeneous expression demands. The layer-based $3/5/7$ setting (CLUIE) provides a better balance between detail preservation and regional structural propagation.

\section{Conclusion}
In this paper, we present CLUIE, a content-adaptive recurrent trajectory modeling framework for underwater image enhancement. Motivated by the spatially heterogeneous nature of underwater degradation, we revisit the sequence organization of visual RWKV and argue that fixed spatial or spectral scans are not always aligned with restoration-related dependencies. To address this issue, we introduce Clustering-aware Semantic Dynamic Reordering, which constructs recurrent token trajectories through semantic clustering and inter-cluster contextual ordering. This allows RWKV states to propagate among regions with closer semantic and degradation-related correlations. To compensate for the local continuity loss introduced by dynamic reordering, we further design Dark-response Modulated Local Propagation, which combines depth-wise convolutional local response extraction with pseudo-dark-response-based modulation. In this way, CLUIE integrates content-adaptive long-range recurrent aggregation with local structural compensation. Experiments on multiple underwater image enhancement benchmarks demonstrate that CLUIE achieves competitive restoration quality and favorable computational efficiency. Ablation studies further verify the contributions of the proposed dynamic reordering and local propagation modules. 

Although CLUIE improves content-adaptive recurrent propagation, it still relies on hard clustering to construct token trajectories. The clustering process introduces additional implementation complexity and may become less stable in extremely low-texture or highly turbid scenes. Moreover, the pseudo-dark response used in DMLP is a feature-space statistic rather than a physical transmission estimate, and its interpretability remains limited. Future work will explore differentiable trajectory learning and physics-guided degradation-aware routing.

\bibliographystyle{IEEEtran}
\bibliography{ref}

\newpage

\end{document}